\documentclass[preprint,12pt]{elsarticle}
\usepackage{amssymb}
\usepackage{booktabs}  %  引入三线表宏包
\usepackage{subfigure}
\usepackage{caption}
\usepackage{multirow}
\usepackage{enumerate}
\usepackage{soul}
\soulregister{\cite}7 % 注册\cite命令
\soulregister{\ref}7 % 注册\ref命令
\soulregister{\pageref}7 % 注册\pageref命令
\journal{Biomedical signal processing and control}

\begin{document}
\begin{frontmatter}
	
\title{Sleep Staging Based on  Multi Scale Dual Attention Network	}
\author[a]{Huafeng Wang\thanks{Corresponding author: wanghuafengbuaa@gmail.com}}
\author[a]{Chonggang Lu}
\author[a]{Qi Zhang}
\author[a]{Zhimin Hu}
\author[b]{Xiaodong Yuan}
\author[b]{Pingshu Zhang}
\author[c]{Wanquan Liu}

\address[a]{School of Information,North China University of Technology, Beijing 100043, China}
\address[b]{Department of Neurology, Kailuan General Hospital, Tangshan, Hebei 063000, China}
\address[c]{School of Intelligent Systems Engineering, Sun Yat-sen University, Shenzhen 518107,  China}
\begin{abstract}
Sleep staging plays an important role on the diagnosis of sleep disorders. In general, experts classify sleep stages manually based on polysomnography (PSG), which is quite time-consuming. Meanwhile, the acquisition process of multiple signals is much complex, which can affect the subject's sleep. Therefore, the use of single-channel electroencephalogram (EEG) for automatic sleep staging has become a popular research topic. In the literature, a large number of sleep staging methods based on single-channel EEG have been proposed with promising results and achieve the preliminary automation of sleep staging. However, the performance for most of these methods in the N1 stage do not satisfy the needs of the diagnosis. In this paper, we propose a deep learning model multi scale dual attention network(MSDAN) based on raw EEG, which  utilizes a one-dimensional convolutional neural network (CNN) to automatically extract features from raw EEG. It uses multi-scale convolution to extract features in different waveforms contained in the EEG signal, connects channel attention and spatial attention mechanisms in series to filter and highlight key information, and uses soft thresholding to remove redundant information. In addition, residual connections are introduced to avoid degradation problems caused by network deepening. Experiments were conducted using two datasets with 5-fold cross-validation and hold-out validation method. The final average accuracy, overall accuracy, macro F1 score and Cohen's Kappa coefficient of the model reach 96.70\%, 91.74\%, 0.8231 and 0.8723 on the Sleep-EDF dataset,  96.14\%, 90.35\%, 0.7945 and 0.8284 on the Sleep-EDFx dataset. Significantly, our model performed superiorly in the N1 stage, with F1 scores of 54.41\% and 52.79\% on the two datasets respectively. The results show the superiority of our network over the existing methods, reaching a new state-of-the-art. In particular, the proposed method achieves excellent results in the N1 sleep stage compared to other methods.
\end{abstract}


\end{frontmatter}

\section{Introduction}
Sleeping is essential for maintaining and regulating various biological functions at the molecular level \cite{laposky_bass_kohsaka_turek_2008}, which helps humans to restore physical and mental health. According to a study by the World Health Organization (WHO) in 2014, nearly one-third of the world's population has a sleep disorder  and 50 million people experience apnoea during sleep\cite{stranges2012sleep}. Chronic sleep disorders have a serious impact on the quality of life and can significantly increase the incidence of cardiovascular disease, diabetes, depression and even cancer, posing a serious threat to human life activities. 

As a consensus, the sleep staging is the most crucial step in sleep analysis. Previously, sleep staging was usually performed manually by an expert based on data recorded on a PSG. The recordings of PSG are usually performed by several sleep experts in sleep clinics. The process generally uses four to six EEG electrodes, two electrooculogram (EOG) electrodes, four electromyogram (EMG) electrodes, two electrocardiogram (ECG) electrodes, and other sensors to collect signals \cite{imtiaz2021systematic}. Many existing studies have had relatively satisfying results using multi-channel signals or a combination of signals for sleep staging. However, the acquisition of multiple signals is complex and can affect the subject's sleep. EEGs record electrical changes in brain activity and contain a wealth of information about the state of the nervous system, which is useful for assessing the health of the brain and diagnosing different sleep and neurological disorders \cite{imtiaz2021systematic}. Therefore, sleep staging based on single-channel EEG becomes the focus of current research on automatic sleep staging methods.

\begin{figure}[htb]
	\centering\includegraphics[width=1\linewidth]{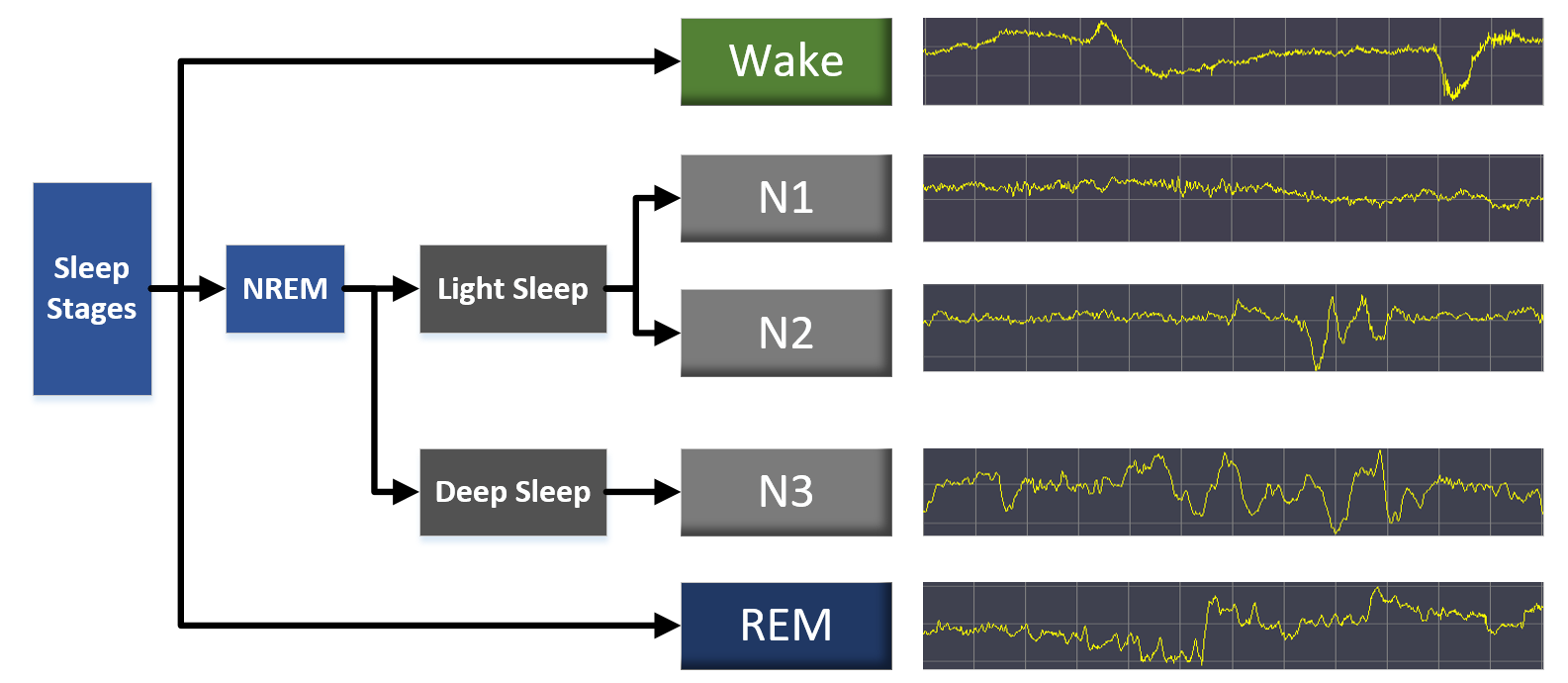}
	\caption{Schematic diagram of the sleep stages}
	\label{Figure01}
\end{figure}

For clinical purposes, experts usually divide sleep into several stages according to PSG. And each stage is characterized by a specific EEG waveform, muscle tension and eye movement \cite{cho2018sleep}. In 1968, the R\&K rules~\cite{wolpert1969manual} suggested to  divide sleep stages into wakefulness, rapid eye movement (REM) and non-rapid eye movement (NREM), while the NREM  was further divided into S1, S2, S3 and S4. Later, American Academy of Sleep Medicine (AASM) \cite{berry2012aasm} modified the R\&K rules by combining S3, S4 into N3 and using N1, N2, N3 as subdivisions of NREM. The relationship between the different sleep stages is shown in Figure~\ref{Figure01}. 

\section{Related work}

As mentioned above, the acquisition of multi-channel signals or multiple signals for sleep staging is definitely complex. Meanwhile, it may cause physical discomfort and mental pressure for the subject. Moreover, the cost to acquire multiple signals is much higher than that of the single signal. Therefore, the current mainstream method focuses on the use of single-channel EEG for automatic sleep staging \cite{chriskos2020review}. In this paper, our proposed dual attention sleep staging model also uses single-channel EEG.

Currently, automatic sleep staging algorithms based on single-channel EEG signals can roughly be categorized into traditional machine learning methods and deep learning-based methods. First of all, traditional  learning models usually rely on handcraft  features. Therefore, a combination of feature extractors and shallow learning models is usually used to construct models. For example, Memar et al.\cite{memar2017novel} extracted 13 features from each sub-epoch decomposed from EEG epoch under different frequencies and  obtained an accuracy of 98.44\%. Tian et al.\cite{tian2017hierarchical} used multiscale entropy as features and  achieved an accuracy of 91.4\%. Lajnef et al.\cite{lajnef2015learning} extracted various features such as variance, kurtosis and permutation entropy from the input signal and  used a multi-class-based support vector machine (SVM) as the classifier to obtain a 92\% accuracy. Hassan et al.\cite{hassan2017automated} applied ensemble empirical mode decomposition (EEMD) to decompose EEG while extracting various features based on statistical moments and  achieved an accuracy of 88.07\%. In summary, all these methods mainly extract features of various manifestations from the raw signal by a feature extractor and then use a single or a combination of multiple classifiers for staging.  And the feature extractors used commonly in EEG processing include short-time Fourier transform (STFT) \cite{samiee2015sleep}, modal decomposition \cite{sharma2017automatic,jiang2019robust}, sparse representation \cite{azadian2019exploiting}, etc., all of which are widely used in those industrial applications. However, the performance of these classical shallow learning models depends heavily on the quality of the features extracted from the acquired signal, i.e., the performance of the feature extractor. In general, the construction of feature extractors often requires relevant a priori knowledge and is applicable to specific data. Nevertheless, the inherent non-linearity and non-stationarity of EEG data and the diversity of individuals make the construction of a feature extractor very time-consuming.

In recent years, as an emerging branch of machine learning, deep learning’s application in all aspects is increasing. Meanwhile, because of its ability to learn high-dimensional and hierarchical features directly from large amounts of data, it is becoming significant in sleep staging. For example, in \cite{jadhav2020automatic}, Jadhav et al. used the wave transform (WT) combined with migration learning to automatically exploit the time-frequency spectrum of EEGs without manual feature extraction, which achieves an accuracy of 83.17\%;  Afterwards, it combined with multiple classifiers to classify the extracted features,  reaching an accuracy of 91.31\%; In \cite{tsinalis2016automatic}, Tsinalis et al. constructed a CNN-based neural network to implement end-to-end learning of raw EEG and finally achieve an average accuracy of 82\% and an overall accuracy of 74\%; In \cite{vilamala2017deep}, VGG-16 is used for spectrogramming of EEG and achieves an accuracy of 86\%; Yang et al.\cite{yang2021single} used one-dimensional CNN to automatically extract features from the raw EEG with HMM for classification which receives an accuracy of 83.98\%; In \cite{seo2020intra}, Seo et al. used a deep CNN-based on a modified ResNet-50 to extract sleep-related features and a two-layer BILSTM to learn the transition rules between sleep stages to achieve end-to-end learning.  Zhang et al. proposed a new unsupervised CNN in \cite{eldele2021attention} to automatically extract sleep-related features from physiological signals and achieved an accuracy of 83.4\%. In one word, deep learning-based methods can greatly facilitate the process of sleep staging without degrading accuracy. Therefore, this method is gradually becoming the focus of current research in the sleep staging field.

However, the acquisition of EEG can be affected by various background noises and individual differences. Meanwhile, in non-stationary states such as sleep, EEGs can be mixed with more complex body movement signals. Thus, deep learning-based sleep staging methods require much deeper one-dimensional structures to extract features of the EEG signals from these complex signals. For example, Yildirim et al.\cite{yildirim2019deep} constructed a one-dimensional CNN-based sleep staging method that exploits the ability of CNN to automatically extract features from the raw PSG, enabling end-to-end learning.  
Humayun et al.\cite{humayun2019end} used a 34-layer deep residual network to build an automatic sleep staging model based on single-channel EEGs and obtained a staging accuracy of 91.4\% on the Sleep-EDFx dataset. Olesen et al.\cite{olesen2018deep} used a 50-layers deep residual network to build an automatic sleep staging model for multiple single-channel PSGs, achieving an accuracy of 84.1\%.  However, the undifferentiated feature extraction is difficult to learn the key information of signal features, that is, salient waveform features, such as alpha rhythm, slow eye movement(SEM), spindle wave, low-amplitude mixed frequency(LAMF),etc. At the same time, several salient waveforms used to distinguish sleep stages generally have different scales, and a single-scale CNN cannot fully capture the salient waveforms features. 

In terms of overall accuracy, the automatic sleep staging methods are currently around 80\% to 90\%. In addition, on the one hand, the overall performance is much difficult to be evaluated because the current researchers usually use different datasets or different versions of the same dataset for experiments, different validation methods, and inconsistent calculations for accuracy. On the other hand, the performance of deep learning methods for the N1 stage is generally poor. For example, according to the confusion matrix in \cite{olesen2018deep}, the F1 score for the N1 stage is 49.7\%. Besides, the calculated F1 scores in \cite{vilamala2017deep} and \cite{tsinalis2016automatic} are 47.3\% and 43.7\% for the N1 stage respectively.  For another instance, the method proposed in \cite{seo2020intra} as the existing state-of-the-art method has an F1 score of only 43.4\% in stage N1. Tsinalis  et al.\cite{tsinalis2016automatic} noted that the poor performance of the N1 stage may be due to the relatively low sleep time in the N1 stage, i.e., less training data. Moreover, Phan et al. \cite{phan2018automatic} claimed that because stage N1 is the transition period between the waking and sleeping states, it contains both $\alpha$ and $\beta$ waves, making it difficult to determine which state affects N1. Therefore, the performance of the current mainstream model in the N1 stage needs further improvement. 
In order to address the above challenges, we propose a multi-scale dual attention network. The main contributions of this paper are as follows:

\begin{itemize}
	\item[-]  A multi scale dual attention network is proposed to conduct sleep staging based on single-channel EEG. It applies CNN to automatically extract the raw single-channel EEG features, and uses the attention mechanisms to capture salient waveform features in the EEG signal, while using multi-scale convolution to adapt to the different scales of salient waveforms.  

\item[-]   An end-to-end method is put forward where features are extracted directly from the raw single-channel EEG without any extra signal processing techniques for staging.

\item[-]  A comprehensive evaluation of the performance of the proposed method is carried out. Comparisons  are conducted with recent studies based on the same single-channel EEG sleep staging to verify the superiority of this method. In addition, the EEGs before and after processing are visualized in a reduced dimension to demonstrate the effectiveness of the method for single-channel EEG feature extraction.

\end{itemize}

\section{The Proposed Method}
\subsection{Model Overview}
\begin{figure}[htb]
	\centering\includegraphics[width=0.9\linewidth]{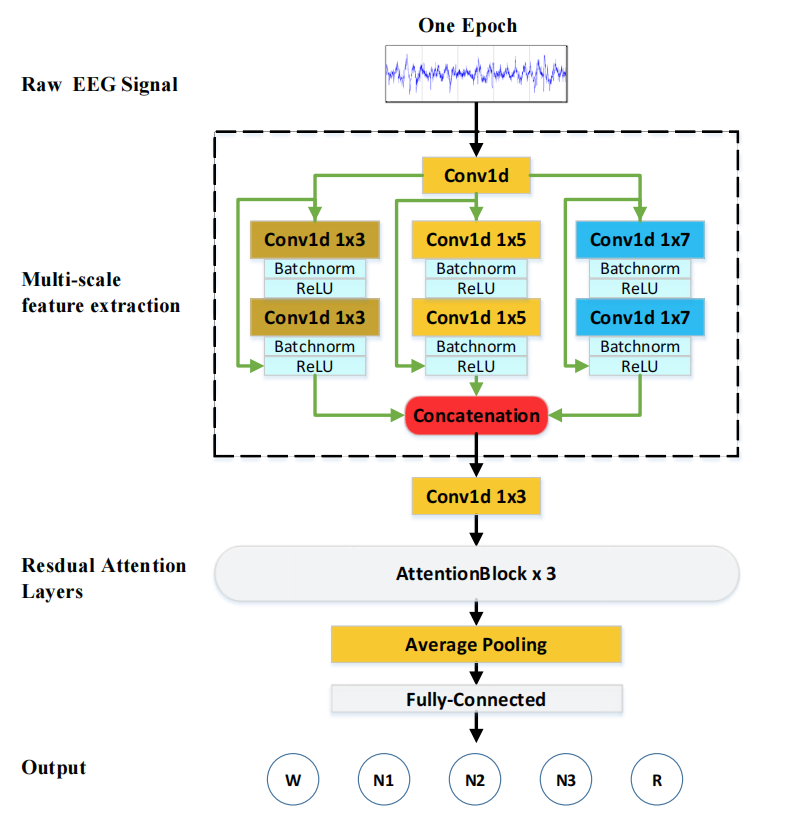}
	\caption{ The architecture of the proposed model.}
	\label{Figure02}
\end{figure}

Figure~\ref{Figure02}  shows an overview of the proposed approach, the new CNN  network is combined with a dual attention mechanism, multi-scale convolution, residual connections and other deep learning optimization techniques to build an end-to-end automatic sleep staging model. The proposed  network consists of three layers. The first layer is a multi-scale convolutional layer with three different branches. This layer uses one-dimensional convolution kernels of different lengths to extract the features of the raw EEG signal. The second layer is the attention  layer, which combines channel attention and spatial attention  to generate more distinguishable feature representations, highlighting important features to improve classification accuracy. In the last layer, the features are sent to the fully connected layer, and finally output is obtained by Softmax. 

The one-dimensional signal of EEG as the input is more likely to cause the problem of the disappearance of the gradient as the network becomes deepen. Luckily, He et al. \cite{he2016deep} proposed the basic residual block which overcomes the problem that the learning rate becomes lower and the accuracy cannot be effectively improved as the network deepening.  Here, we introduce residual connections for the multi-scale layer and the attention  layer in the network to overcome the degradation problem in the traditional deep network. 

\subsection{Multi-scale feature extraction}
 Multi-scale feature extraction module is composed of three branches with different one-dimensional convolution kernel sizes, which are $1\times3$, $1\times5$, and $1\times7$. Each branch includes two convolutional layers, batch normalization, activation function ReLU and residual connection. The outputs of the three branches are combined through a concatenate operation. These concatenated features are then forwarded to the Resdual Attention Layers. For the judgment of a certain sleep stage, it is sometimes necessary to combine multiple waveforms. Using multi-scale convolution, it is possible to capture prominent waveforms while focusing on the overall trend. For example, in the N1 period, while focusing on low-amplitude mixing frequencies, details such as slow eye movements can be captured.

\subsection{Residual Attention Layer }
\begin{figure}[htbp]
	
	\centering\includegraphics[width=1\linewidth]{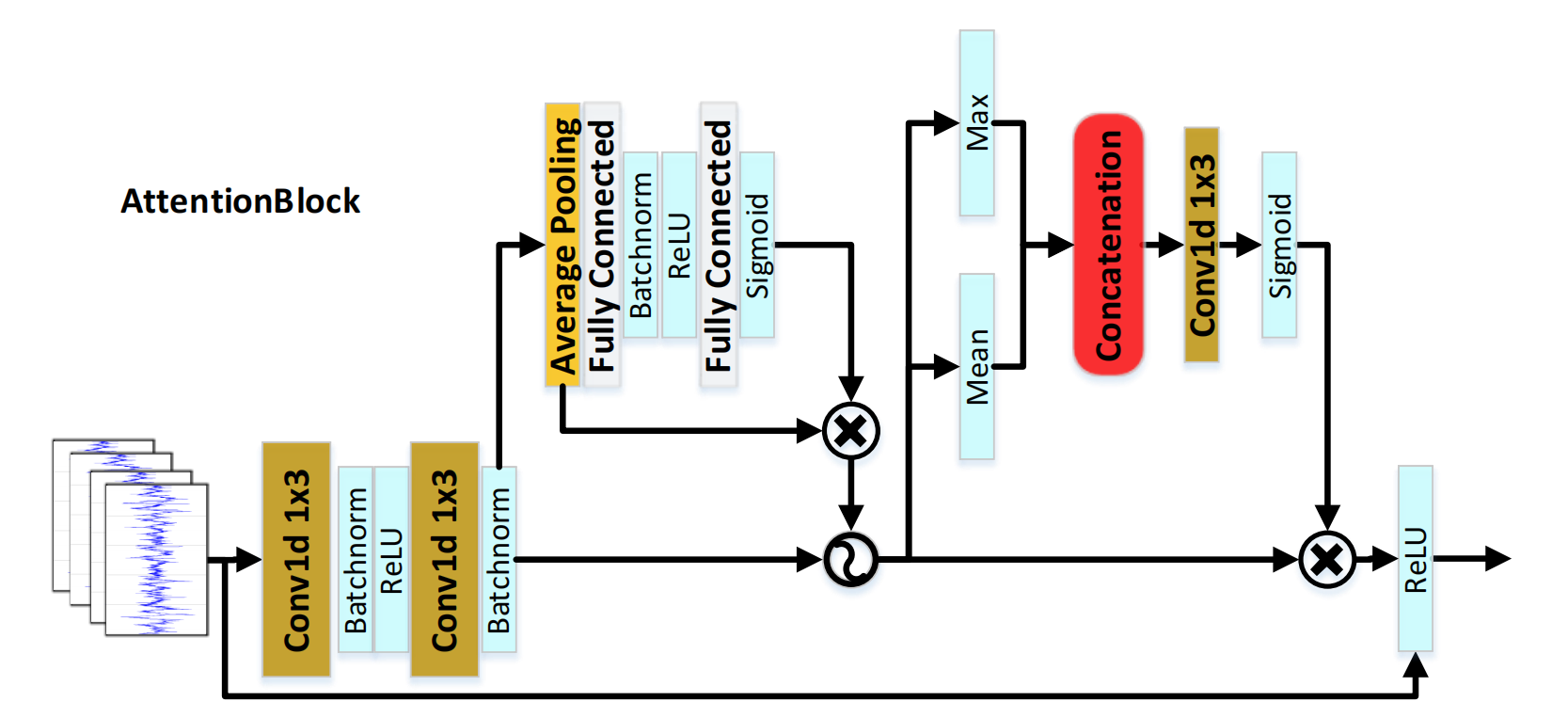}
	\caption{AttentionBlock}
	\label{Figure03}
\end{figure}

 As shown in Figure~\ref{Figure02},  the residual attention Layer  contains  three AttentionBlocks in series. The AttentionBlock shown in Figure~\ref{Figure03} is mainly composed of two one-dimensional convolutionnal layers, a channel attention part and a spatial attention part, which are connected in one pipeline. 
	
 The attention mechanism in deep learning is essentially similar to the human selective visual attention mechanism, which can select information that is more critical to the task  from a large number of information. Donoho et al. \cite{donoho1995noising} proposed a very simple soft threshold method for denoising. It makes the setting of the value intervals for feature more flexible compared to the ReLU function. In \cite{zhao2019deep}, Zhao et al. inserted soft thresholding into the deep architectures for fault diagnosis. The EEG signal is a signal with a lot of noise, so it is of great significance to use the attention mechanism combined with soft threshold to eliminate redundant information. In the AttentionBlock proposed in this paper, the channel attention part and the spatial attention part are connected in series for sleep staging, at the same time, the soft threshold is introduced in the channel attention part to eliminate unimportant features in advance. Therefore, the model can more effectively extract channel features and one-dimensional spatial features in EEG signals that are more conducive to identifying different sleep stages.

The  channel attention uses the weights in combination with the soft threshold function to optimize the channel features and then obtain the refined channel features. Afterwards, the optimized channel features are then fed into the spatial attention model to extract key information from the one-dimensional signal to obtain improved semantic information. Finally, the improved semantic information is connected to the identical mapping to form a residual attention block. 

\subsubsection{Channel attention mechanism}

A one-dimensional matrix $P\in R^{C\times1}$ composed by different weights is learned for each channel through channel attention. Afterwards, the channel attention applies multiplication to focus on the channel containing more meaningful information. For a given multi-channel feature signal $F\in R^{C\times W}$ (where C and W respectively represent the number and width of the signal channel), a set of weights is formed after the channel attention module, denoted as $F^{\prime}\in R^{C\times1}$.

\begin{figure}[htb]
	
	\centering\includegraphics[width=0.5\linewidth]{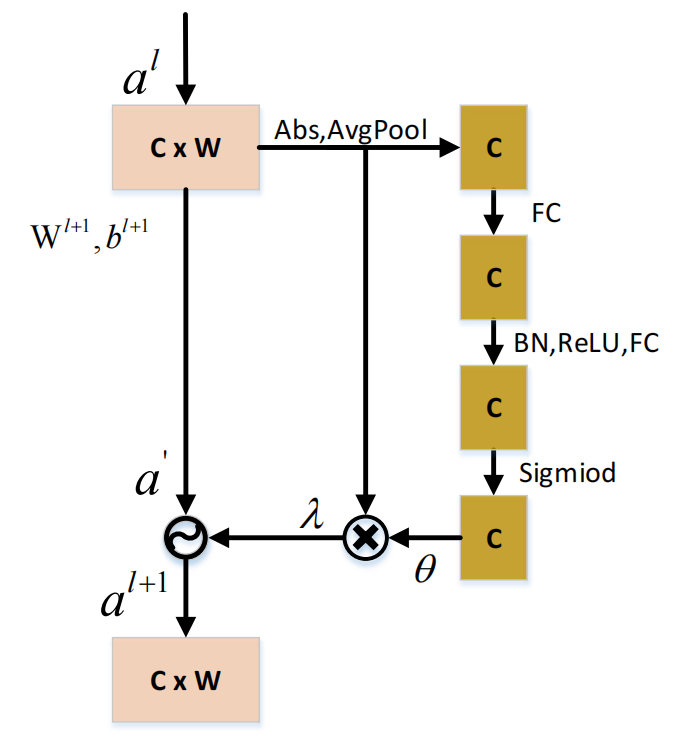}
	\caption{ Channel attention}
	\label{Figure04}
\end{figure}

As shown in Figure~\ref{Figure04}, the input $a^{l}$ learns a set of thresholds $\lambda$ with a value interval between 0 and 1 through the constructed sub-network. The obtained thresholds are then used to soft-threshold the features to obtain $a^{l+1}$. In this network, the thresholds are adjustable automatically according to the samples themselves through an attention mechanism. The output of each layer is as follows,

\begin{equation}
a^{\prime}=w^{l+1}a^{l}+b^{l+1}
\end{equation}

\begin{equation}
a^{l+1}=soft(a^{\prime},\theta\lambda)=sgn(a^{\prime})( |a^{\prime}|-\theta\lambda)_{+}
\end{equation}
where soft refers to the soft thresholding, $\theta$ refers to a set of weights learned  through the channel attention mechanism, $\lambda$ refers to the obtained threshold, and sgn refers to the signum function, $W^{l+1}$ refers to the weight matrix between $a^{l}$ and $a^{\prime}$, $b^{l+1}$ refers to the bias vector corresponding to $W^{l+1}$.

\subsubsection{Spatial attention mechanism}

\begin{figure}[htb]
	
	\centering\includegraphics[width=0.5\linewidth]{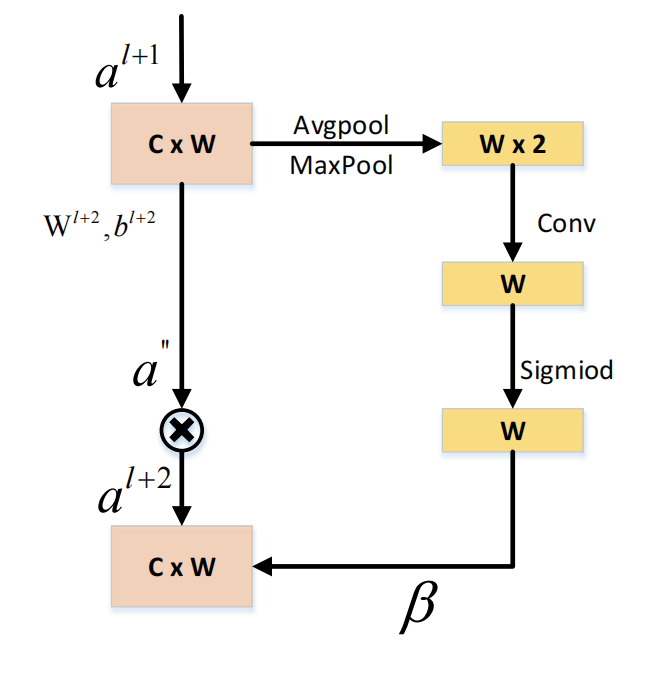}
	\caption{ Spatial attention}
	\label{Figure05}
\end{figure}

Spatial attention aims to discover the relationships between spatial locations. As shown in Figure~\ref{Figure05}, for spatial attention, different feature descriptions are first generated for each spatial location using global average pooling and max pooling. Afterwards, the spatial attention features are obtained using one-dimensional convolution  and finally scaled to 0 to 1 using the Sigmoid function. The feature $a^{l+1}$ obtained by channel attention becomes $a^{l+2}$ after being refined by the spatial attention mechanism. The output of each layer is as follows,

\begin{equation}
\beta=\delta(Conv^{1\times3}(AvgPool(a^{l+1});Maxpool(a^{l+1})))
\end{equation}

\begin{equation}
a^{\prime\prime}=w^{l+2}a^{l+1}+b^{l+2}
\end{equation}

\begin{equation}
a^{l+2}=a^{\prime\prime}*\beta
\end{equation}
where Avgpool and MaxPool are average pooling and max pooling respectively, $Conv^{1\times3}$ is a convolution operation with a convolution kernel of 1$\times$3 ,  $\delta$ is Sigmoid activation function, $\beta$ refers to a set of weights learned through the spatial attention mechanism, $W^{l+2}$ refers to the weight matrix between $a^{l+1}$ and $a^{\prime\prime}$, $b^{l+2}$ refers to the bias vector corresponding to $W^{l+2}$.

 In summary, we propose an end-to-end deep CNN for sleep staging, for one-dimensional EEG signals, our model utilizes multi-scale convolution to extract rich information from the original signal, and uses the attention mechanism to filter out redundant information to capture salient waveform features used to distinguish different sleep stages. 
\subsection{  Experimental Configuration }

This study uses Softmax to preprocess the input. Softmax is widely used in multi-stage tasks and is defined as follows:
Assuming that the input array x contains j elements, $x_{i}$ representing the i-th element in x, then the Softmax value $y_{i}$ of this element is:
\begin{equation}
y_{i}=softmax(x_{i})=\frac{exp(x_{i})}{\sum_{j}exp(x_{j})}
\end{equation}
As can be seen from Table~\ref{Table01}, sleep staging is a staging task with an unbalanced number of labels in each category, so the difference in the amount of data across labels needs to be balanced. The model in this paper uses standard multi-class cross entropy as the loss function, which is defined as follows:
\begin{equation}
loss(x,class)=-\log(\frac{exp(x[class])}{\sum_{j}exp(x[j])})=-x[class]+\log(\sum_{j}exp(x[j])) 
\end{equation}
where x is the prediction vector output by the model for a certain epoch, class represents the index of the actual label corresponding to the epoch, j is the number of classes, x[class] represents the element whose index is class in the input tensor. In order to further solve the problem of label imbalance during sleep, we add a weight determined by the proportion of labels to the loss function and limit the value of weight to between 1 and 5:
\begin{equation}
loss(x,class)=weight[class]\times(-x[class]+\log(\sum_{j}exp(x[j])) )
\end{equation}
\begin{equation}
weight[class]=min(5,max(1,\ln(\frac{1}{p(class)})))
\end{equation}
where weight[class] is the weight of the actual label, p(target) is the proportion of a certain category label to the total label. 

For a batch containing N epochs, take the average value as the final loss:
\begin{equation}
loss=\frac{\sum_{i=1}^{N}loss(i,class[i])}{\sum_{i=1}^{N}weight[class[i]]}
\end{equation}
According to the loss function, the Adam optimization method was used for optimization, with the learning rate set to 0.0005 and the batch size to 8.

\section{Experiment and analysis}

\subsection{EEG datasets}

We use two public datasets from PhysioBank \cite{goldberger2000physiobank}: Sleep-EDF database and Sleep-EDFx database [expanded]\cite{kemp2000analysis}. The two datasets were chosen to provide a better comparison with known studies as a large amount of existing research was conducted on these two datasets.

The Sleep-EDF dataset is recorded from eight white men and women aged 21-35 years and contains horizontal EOGs sampled at 100 Hz and EEGs from the Fpz-Cz and Pz-Oz channel. These eight recordings were divided into two subsets, named sc* and st*: sc* contains sub-chin EMG envelopes, oronasal airflow, rectal temperature and event markers sampled at 1 Hz; st* contains sub-chin EMG sampled at 100 Hz and event markers sampled at 1 Hz. Sleep stages are classified according to the R \& K rules \cite{wolpert1969manual} and include W, 1, 2, 3, 4, R, M and 'unscored'. In this paper M and 'unscored', which account for a very small proportion, are removed, while S3 and S4 are combined into N3 according to the latest AASM standards \cite{berry2012aasm}.

The Sleep-EDFx dataset contains 197 complete PSG records with EEG, EOG, chin EMG and event markers. All processing cases are the same as the Sleep-EDF dataset.

The sleep stage statistics for the Sleep-EDF and Sleep-EDFx datasets are shown in Table~\ref{Table01}.

\begin{table}[htbp]
	\centering
	\caption{The number and proportion of sleep stages in the two datasets}
	\resizebox{1\columnwidth}{!}{
		\begin{tabular}{c c c c c c c}
			\toprule
			\textbf{ } & \textbf{W} & \textbf{N1} & \textbf{N2} & \textbf{N3} & \textbf{R} & \textbf{Total}\\
			\midrule
			Sleep-EDF & 8030(52.8\%) & 604(4.0\%) & 3621(23.8\%) & 1299(8.5\%) & 1609(10.6\%) & 15199(100\%) \\
			\midrule 
			Sleep-EDFx & 289665(63.3\%) & 25174(5.5\%) & 88983(19.4\%) & 19454(4.3\%) & 34184(7.5\%) & 238976(100\%) \\
			\bottomrule  % 底部线
	\end{tabular}}
	\label{Table01}
\end{table}

\begin{figure}[htbp]
	\centering\includegraphics[width=1\linewidth]{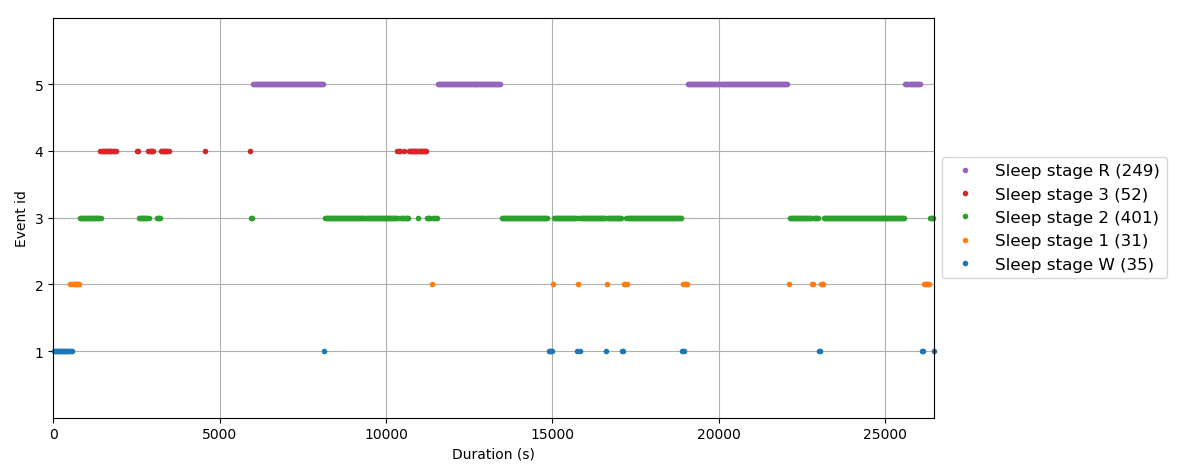}
	\caption{The stages of sleep throughout the night}
	\label{Figure06}
\end{figure}

Figure~\ref{Figure06} shows the change in sleep stages throughout the night for a subject in the Sleep-EDFx dataset, which source from ST7142J0-PSG.edf\cite{kemp2000analysis}. The subject experienced 249 REM stages, 31 N1 stages, 401 N2 stages, 52 N3 stages and 35 W stages during the night.

\subsection{Prerocessing}

Traditional data pre-processing methods for EEGs require low-pass filtering at the upper frequency of $\beta$ (e.g., 30Hz) because EEGs contain mostly valid signals smaller than 30Hz. However, for CNNs, it can automatically extract the valid information without performing filtering. In order to accommodate different PSG acquisition devices and individual differences between subjects, the EEGs of each subject were normalized as described in~\cite{vilamala2017deep}, i.e.,

\begin{equation}
x_{norm}=2\frac{x-S_{0.05}(x)}{S_{0.95}(x)-S_{0.05}(x)}-1
\end{equation}
where $x_{norm}$ is the raw signal, $S_{0.95}$ is the signal at the 95th quantile and $S_{0.05}$ is the signal at the 5th quantile, with processed data point values in the range [- 1,1]. Each input data is flipped with a probability of 0.5 and added a random noise of 1\% is to increase it. Each 30-second segment of the data corresponds to a sleep stage. In order to input the data into the network, the sleep stages were numbered and mapped according to the relationship in equation (2) as follows:

\begin{equation}
N3\rightarrow0,  N2\rightarrow1,  N1\rightarrow2,  R\rightarrow3,  W\rightarrow4 
\end{equation}

\subsection{The methods of evaluation}

In this paper, two cross-validation methods are used to evaluate the system performance, named k-fold cross-validation and hold-out validation. For k-fold cross-validation, the whole dataset is randomly divided into k validation sets. Afterwards, the k validation sets are divided into epoch as the smallest unit, while the remaining k-1 subsets are used as the training set. Subsequently, k experiments are conducted and the results of all validation sets are averaged to obtain the final results. In this paper, k = 5 is chosen and the system performance is evaluated using the two datasets. For the hold-out validation, the dataset is divided into a training set and a validation set. The ratio of 8:2 is used to divide the training set and the validation set. The hold-out method can effectively demonstrate the generalization ability of the system to unknown data (subjects), which has high practical application value. In addition, under the fact that the sample size of the Sleep-EDF dataset is too small, it is not used for hold-out validation in this paper.

Hold-out validation: The division of the training set and the validation set in the Sleep-EDFx data set is shown in Table~\ref{Table02}.

\begin{table}[htbp]
	\centering
	\caption{Sleep-EDFx training set validation set division}
	\resizebox{1\columnwidth}{!}{
		\begin{tabular}{c c c c c c c}
			\toprule
			\textbf{ } & \textbf{Number of subjects} & \textbf{W} & \textbf{N1} & \textbf{N2} & \textbf{N3} & \textbf{R} \\
			\midrule
			Total & 197 & 289665(63.3\%) & 25174(5.5\%) & 88983(19.4\%) & 19454(4.3\%) & 34184(7.5\%) \\
			\midrule 
			Train & 157 & 232684 & 19553 & 70711 & 15960 & 27060 \\
			\midrule
			Eval & 40 & 56981 & 5621 & 18272 & 3494 & 7124 \\
			\bottomrule  % 底部线
	\end{tabular}}
	\label{Table02}
\end{table}

Moreover, a variety of metrics are used to evaluate the performance of the system. When it comes to a multi-class problem, each stage is evaluated by treating the stage as a positive example and the other stages as negative examples. For each sleep stage, four metrics are used for evaluation, which are accuracy, recall rate, precision, and F1 score respectively. The metrics for each stage will be averaged as an overall metric. The metrics are expressed as follows:		

\begin{equation}
Accuracy(acc)=\frac{TP+TN}{TP+TN+FP+FN}(\%)
\end{equation}
\begin{equation}
Recall(rec)=\frac{TP}{TP+FN}(\%)
\end{equation}
\begin{equation}
Precision(pr)=\frac{TP}{TP+FP}(\%)
\end{equation}
\begin{equation}
F1\ Score(F1)=\frac{2TP}{2TP+FP+FN}(\%)
\end{equation}

Meanwhile, to evaluate the overall performance of the system, the overall accuracy and the Cohen's kappa coefficient will be calculated as follows:		

\begin{equation}
Overall\ Accuracy=\frac{\sum(1|Y_{i}= PY_{i})}{\sum(1|Y_{i}=Y_{i})}(\%)
\end{equation}
\begin{equation}
Cohen's\ kappa =\frac{p_{0}-p_{e}}{1-p_{e}},p_{e}=\frac{\sum a_{i}b_{i}}{n^{2}},p_{0}=Overall\ accuracy
\end{equation}
where $Y_{i}$ denotes the true label of the $i^{th}$ sample, $PY_{i}$ denotes the model predicted label of the $i^{th}$ sample, $a_{i}$ is the number of the $i^{th}$ actual sample, $b_{i}$ is the number of the $i^{th}$ predicted sample, and $n$ is the total number of samples. Kappa coefficient is used to measure the degree of coincidence between the model staging result and the real result. the closer the value of kappa coefficient is to 1, the milder the two, which illustrates the better the performance of the model.

We use the PyTorch 1.5 framework to implement the model, running on an Ubuntu 16.04 system with TiTan X GPU, Intel Xeon E5 CPU and 54.9G RAM.

\subsection{Experimental results of a dual attention network}

%			\multicolumn{10}{c}{Confusion matrix and each evaluation indicator}  \\
\begin{table}[htb]
	\centering
	\caption{Confusion matrix and performance metrics in the Sleep-EDF dataset validation experiment with the 5-fold method}
	\resizebox{1\columnwidth}{!}{
		\begin{tabular}{c c c c c c c c c c}
			\toprule
			& N3 & N2 & N1 & R & W & Acc(\%) & Re(\%) & Pr(\%) & F1(\%) \\
			\midrule
			W & 9 & 2 & 97 & 58 & 7792 & 98.44 & 97.91 & 99.12 & 98.51 \\
			\midrule
			R & 1 & 111 & 153 & 1310 & 20 & 96.28 & 82.13 & 82.70 & 82.42 \\
			\midrule
			N1 & 6 & 71 & 364 & 123 & 38 & 95.96 & 60.36 & 49.52 & 54.41 \\
			\midrule
			N2 & 185 & 3198 & 118 & 91 & 8 & 95.15 & 88.83 & 90.65 & 89.73 \\
			\midrule
			N3 & 1134 & 144 & 2 & 2 & 3 & 97.66 & 88.25 & 84.75 & 86.47 \\
			\bottomrule  % 底部线
	\end{tabular}}
	\label{Table03}
\end{table}

%\multicolumn{10}{c}{Confusion matrix and each evaluation indicator}  \\
\begin{table}[htb]

	\centering
	\caption{Confusion matrix and performance metrics in the Sleep-EDFx dataset validation experiment with the hold-out method}
	\resizebox{1\columnwidth}{!}{
		\begin{tabular}{c c c c c c c c c c}
			\toprule		
			& N3 & N2 & N1 & R & W & Acc(\%) & Re(\%) & Pr(\%) & F1(\%) \\
			\midrule
			W & 12 & 48 & 1438 & 363 & 55132 & 97.04 & 96.73 & 98.49 & 97.60 \\
			\midrule
			R & 9 & 665 & 1006 & 5343 & 104 & 96.08 & 74.97 & 74.74 & 74.85 \\
			\midrule
			N1 & 83 & 1055 & 3041 & 793 & 653 & 92.39 & 54.05 & 40.97 & 46.61 \\
			\midrule
			N2 & 1321 & 14299 & 1918 & 649 & 86 & 93.28 & 78.25 & 86.78 & 82.30 \\
			\midrule
			N3 & 3072 & 408 & 19 & 1 & 2 & 97.97 & 87.72 & 68.27 & 76.78 \\
			\bottomrule  % 底部线
	\end{tabular}}
	\label{Table04}
\end{table}

\begin{table}[h]

	\centering
	\caption{Confusion matrix and performance metrics in the Sleep-EDFx dataset validation experiment with the 5-fold method}
	\resizebox{1\columnwidth}{!}{
		\begin{tabular}{c c c c c c c c c c}
	\midrule
& N3 & N2 & N1 & R & W & Acc(\%) & Re(\%) & Pr(\%) & F1(\%) \\
			\midrule
			W & 97 & 379 & 7807 & 1105 & 280164 & 97.51 & 96.76 & 99.29 & 98.01 \\
			\midrule
			R & 25 & 2196 & 4078 & 27578 & 293 & 96.99 & 80.71 & 79.33 & 80.01 \\
			\midrule
			N1 & 93 & 4477 & 16384 & 2802 & 1401 & 93.59 & 65.12 & 44.38 & 52.79 \\
			\midrule
			N2 & 4752 & 72114 & 8543 & 3258 & 284 & 94.25 & 81.07 & 88.43 & 84.59 \\
			\midrule
			N3 & 16917 & 2385 & 104 & 21 & 23 & 98.36 & 86.98 & 77.30 & 81.85 \\
		
			\bottomrule  % 底部线
	\end{tabular}}
	\label{Table05}
\end{table}

\begin{table}[h]

	\centering
	\caption{ Overall performance indicators}
	\resizebox{1\columnwidth}{!}{
		\begin{tabular}{c c c c c c}
			\toprule
			Overall & Re(\%) & Mean Acc & Overall Acc & Kappa & MF1 \\
			\midrule
			Sleep-edf\\(5-fold) & 83.50 & 96.70 & 91.74 & 0.8723 & 0.8231 \\
			\midrule
			Sleep-edfx\\(hold-out) & 78.35 & 95.35 & 88.38 & 0.7963 & 0.7563 \\
			\midrule
			Sleep-edfx\\(5-fold) & 82.13 & 96.14 & 90.35 & 0.8284 & 0.7945 \\
			\bottomrule  % 底部线
	\end{tabular}
}
	\label{Table06}
\end{table}

Experimental results in Table~\ref{Table03} show the confusion matrix results and performance metrics for each sleep stage obtained for Sleep-EDF after using the 5-fold method. Table~\ref{Table04} and~\ref{Table05} show the confusion matrix results and performance metrics results for each sleep stage obtained for Sleep-EDFx after the hold-out and 5-fold validation respectively. Table~\ref{Table06} shows the overall performance metrics of the model on both datasets.

The experimental results of the 5-fold method show that the overall recall rate, average accuracy, overall accuracy, kappa coefficient and macro F1 score of the staging obtained in the Sleep-EDF dataset are 83.50\%, 96.70\%, 91.74\%, 0.8723 and 0.8231 respectively. For each sleep stage, the F1 score is 98.51\% in the W period, and the score in the N1 period is the lowest, which is 54.41\%. The staging overall recall rate, average accuracy, overall accuracy, kappa coefficient and macro F1 score obtained in the Sleep-EDFx dataset are 82.13\%, 96.14\%, 90.35\%, 0.8284 and 0.7945 respectively. The F1 score of the N1 stage is 52.79\%. Compared with the Sleep-EDF dataset, the overall performance of the system is degraded mainly due to the substantial increase in the amount of data, which makes the network unable to fully learn all the features.

\begin{figure}[htbp]

	\centering\includegraphics[width=1\linewidth]{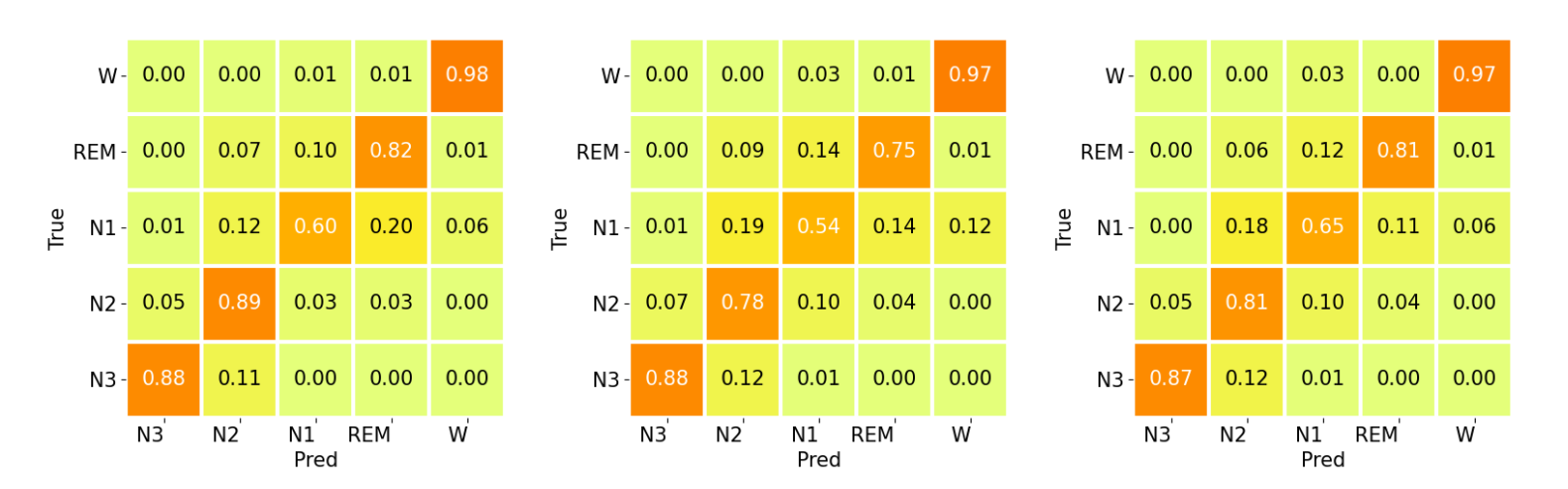}
	\caption{Visualisation of the experimental confusion matrix for: Sleep-EDF 5-fold validation, Sleep-EDFx 5-fold validation, Sleep-EDFx hold-out method validation}
		\label{Figure07}
\end{figure}

Preliminary results of the experiment show that the network proposed in this paper can provide better staging performance for both datasets. However, it can be found that the network has a higher accuracy for the W stage and the worst accuracy for the N1 stage among the stages. Further validation is needed to examine if this method is an improvement towards other methods. In particular, the confusion matrix analysis reveals that the N1 stage can be easily misclassified as the N2 and R stages. Figure~\ref{Figure07} represents the visualisation of the confusion matrix in experimental results. The meaning of each square is the proportion of the number of that stage to the actual number of total labels. The darker the colour, the higher the proportion of that stage. Meanwhile, the figure shows that the correct rate of the W stage is high and the correct rate of the N1 stage is generally low.

\subsection{Comparison experiments}

 In order to conduct a further comparison, this paper selects some methods of using Sleep-EDFx and compares their overall performance and performance in the N1 stage. As shown in Table~\ref{Table07}, the data division and verification methods used in different literature are not completely the same. Verification methods include hold-out validation \cite{jadhav2020automatic,olesen2018deep}, k-fold cross-validation \cite{tsinalis2016automatic,perslev2019u} and leave-one-out cross-validation (LOO-CV) \cite{phan2018automatic,supratak2017deepsleepnet}. In the meantime, the calculation methods of the overall accuracy are inconsistent. For example, in \cite{memar2017novel,yang2021single}, the average accuracy of each sleep stage is taken as the final accuracy; in \cite{zhang2016automatic}, the weighted average accuracy of each sleep stage is taken as the final accuracy. In order to make a fair comparison, in Table~\ref{Table07}, according to the confusion matrix given by the authors and the calculation methods used in this paper, the overall accuracy and F1 score of the methods listed are recalculated. According to Table~\ref{Table07}, the overall accuracy of the hold-out validation and the k-fold cross-validation used in this paper reach 88.38\% and 90.35\%. The F1 score in the N1 stage even reaches 52.8\%, which proves that our model has excellent overall performance, especially in the N1 Stage.
 
 \begin{table}[htbp]

 	\centering
 	\caption{ Comparison with mainstream methods}
 	\resizebox{1\columnwidth}{!}{
 		\begin{tabular}{c c c c c c c}
 			\toprule
 			Method & Channel & Split & Acc(overall)(\%) & Kappa & MF1(\%) & F1(N1)(\%) \\
 			\midrule
 			CNN \cite{tsinalis2016automatic} & Fpz-Cz & 20-fold CV  & 74.8 & - & 81.0 & 43.67 \\
 			DeepSleepNet \cite{supratak2017deepsleepnet} & Fpz-Cz & 20 subjects LOO-CV & 82.0 & 0.76 & 76.9 & 46.6\\
 			CWT+ SqueezeNet \cite{jadhav2020automatic}
 			 & Fpz-Cz &  training-validation-testing
 			70\%-10\% -20\% & 83.2 & - & - & 34.5 \\
 			Utime \cite{perslev2019u} & Fpz-Cz & 5-fold CV & 81.3 & - & 76.0 & 51.0 \\
 			BLSTM-SVM \cite{phan2018automatic} & Fpz-Cz & 20 subjects LOO-CV & 82.5 & 0.76 & 72.0 & 23.8 \\
 			ResNet 34 \cite{humayun2019end} & Fpz-Cz & 70\%-30\% subjects & 91.4 & - & - & - \\
 			ResNet 50 \cite{olesen2018deep} & Fpz-Cz &  training-validation-testing
 			80\%-10\% -10\% & 84.1 & 0.746 & 83.1 & 49.7 \\
 			FT SeqSleepNet+ \cite{phan2020towards}  & Fpz-Cz & 20subjects LOO-CV & 85.2 & 0.789 & 79.6 & 50.9 \\
 			Multitask 1-max CNN \cite{phan2018joint} & Fpz-Cz & 20subjects LOO-CV & 81.9 & 0.74 & 73.8 & 40.5 \\
 			IITNet \cite{seo2020intra} & Fpz-Cz & 20-fold CV subjects & 83.9 & 0.78 & 77.6 & 43.4 \\
 			CNN-HMM \cite{yang2021single} & Fpz-Cz & 20-fold CV subjects & 83.98 & 0.78 & 76.9 & 35.1 \\
 			MRCNN-TCE \cite{eldele2021attention} & Fpz-Cz & 20-fold CV subjects & 82.9 & 0.77 & 78.1 & 47.4\\
 			Proposed Method & Fpz-Cz & training-validation
 			80\%-20\%  & 88.38 & 0.7963 & 75.63 & 46.6 \\
 			Proposed Method & Fpz-Cz & 5-fold CV & 90.35 & 0.8284 & 79.45 & 52.8 \\
			
 			\bottomrule  % 底部线
 	\end{tabular}}
  	\label{Table07}
 \end{table}

In the literature, because the training and validation methods used are inconsistent, and most of them do not give confusion matrices for the experimental results, it is one-sided if simply quoting the experimental results from the original literature. Therefore, to more reasonably validate the staging performance of the models, further experiments were conducted with ResNet50 \cite{olesen2018deep}, ResNet34 \cite{humayun2019end}, CNN \cite{tsinalis2016automatic}, Utime \cite{perslev2019u} under the same conditions. Experiments with all models should meet the following requirements:

(1) All methods should use the data of 8 subjects in Sleep-EDF, the Fpz-Cz channel and 5-fold cross-validation to validate.

(2) The number of training sessions and batch size should be the same. The batch size here is 64.

(3) All data should be pre-processed as described in the "pre-processing of data" section of this paper. 

(4) To accommodate the EEG, the convolution operation should be one-dimensional in all models.

\begin{table}[htbp]
	
	\centering
	\caption{ Results of the five methods at stage N1 and overall assessment metrics}
	\resizebox{1\columnwidth}{!}{
		\begin{tabular}{cccccc}
			\toprule
			Method & N1/Overall & Re(\%) & Pr(\%) & F1(\%) & Acc(overall) \\
			\midrule
			\multirow{2}*{Ours} & Overall & 84.37 & 80.22 & 81.77 & \multirow{2}*{91.05} \\
			~ & N1 & 67.55 & 43.99 & \textbf{53.28} \\
			\midrule
			\multirow{2}*{ResNet50\cite{olesen2018deep}} & Overall & 70.91& 67.19 & 68.60 & \multirow{2}*{88.63} \\
			~ & N1 & 61.87 & 41.20 & \textbf{49.47} \\
			\midrule
			\multirow{2}*{ResNet34\cite{humayun2019end}} & Overall & 82.12 & 78.43 & 79.64 & \multirow{2}*{89.97} \\
			~ & N1 & 65.05 & 39.78 & \textbf{49.37} \\
			\midrule
			\multirow{2}*{CNN\cite{tsinalis2016automatic}} & Overall & 83.23 & 77.77 & 79.67 & \multirow{2}*{89.61}\\
			~ & N1 & 68.73 & 40.69 & \textbf{51.12} \\
			\midrule
			\multirow{2}*{Utime\cite{perslev2019u}} & Overall & 76.76 & 75.38 & 76.04 & \multirow{2}*{88.27}\\
			~ & N1 & 43.09 & 43.02 & \textbf{43.05} \\
			\bottomrule  % 底部线
	\end{tabular}}
	\label{Table08}
\end{table}

\begin{figure}[htbp]

	\centering\includegraphics[width=0.8\linewidth]{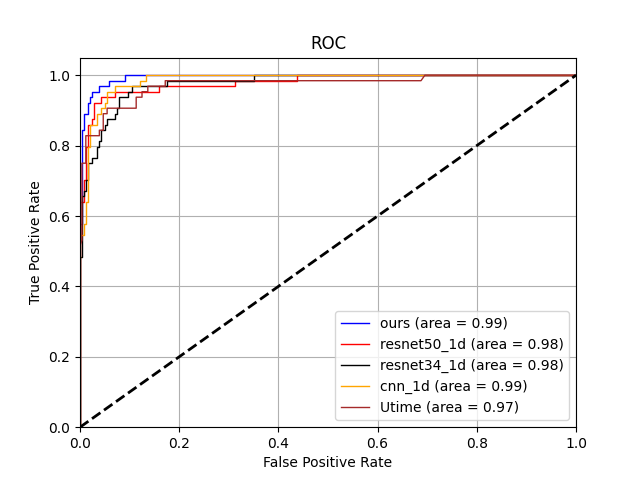}
	\caption{The ROC curves in the comparison experiment}
		\label{Figure08}
\end{figure}

\begin{figure}[htbp]

	\centering\includegraphics[width=0.8\linewidth]{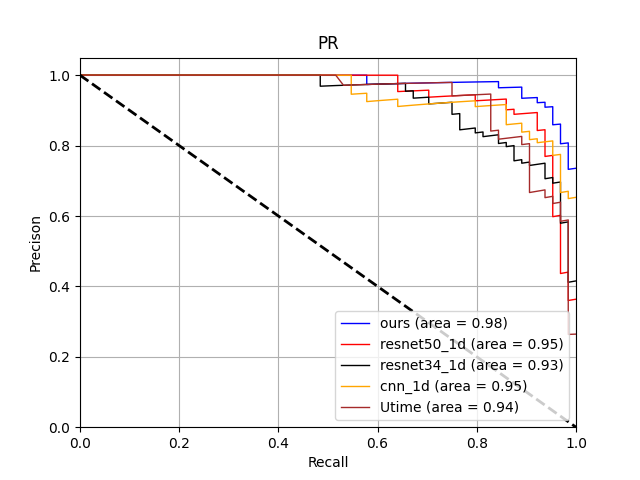}
	\caption{The PR curves in the comparison experiment}
		\label{Figure09}
\end{figure}

Figure~\ref{Figure08} and Figure~\ref{Figure09} shows the ROC and PR curves of the comparison experiment, and Table~\ref{Table08} shows the overall performance metrics of each method. ROC curve and PR curve show that the performance of the network model in this paper is better than other network models. Meanwhile, the data in Table~\ref{Table07} shows that our method is better than other methods in terms of overall accuracy, overall recall, overall precision and F1 score. Moreover, our method also shows its superiority in the N1 stage. The F1 score of the method in the N1 stage is improved by nearly 4\%, 4\%, 2\% and 10\% compared with Resnet50, Resnet34, CNN and Utime respectively. In summary, the network model and method in this paper can completely improve the overall performance of the model, and it has a good effect in the N1 stage.

\section{Discussion and analysis}

\subsection{Comparison with manual staging}

\begin{figure}[htb]

	\centering\includegraphics[width=1\linewidth]{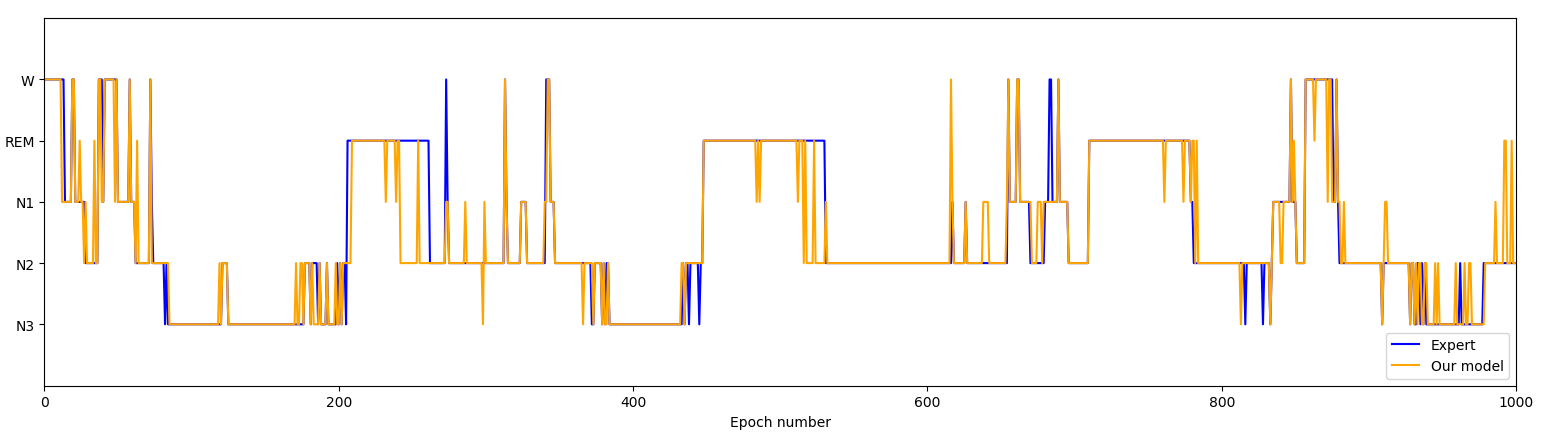}
	\caption{Comparison of manual staging (blue) and automatic staging by the proposed method (orange)}
		\label{Figure10}
\end{figure}

In order to compare with the performance of the manual staging model, the model in this paper trained by the Sleep-EDFx dataset is classified automatically for the experiment. The visualisation results are shown in Figure~\ref{Figure10}. Silbe et al.~\cite{silber2007visual} gives a standard range of accuracy of manual staging which is W (68-89\%), N1 (23-74\%), N2 (79-90\%), N3 (45-80\%) and R (78-94\%). Combined with the results in Table~\ref{Table07}, it can be seen that the proposed model meets the requirements of each criterion and reaches the level of practical application.

\subsection{Visual network for sleep EEG feature extraction}

\begin{figure}[htbp]

	\subfigure[Before] %第一张子图
	{
		\begin{minipage}[t]{0.5\linewidth}
			\centering          %子图居中
			\includegraphics[scale=0.85]{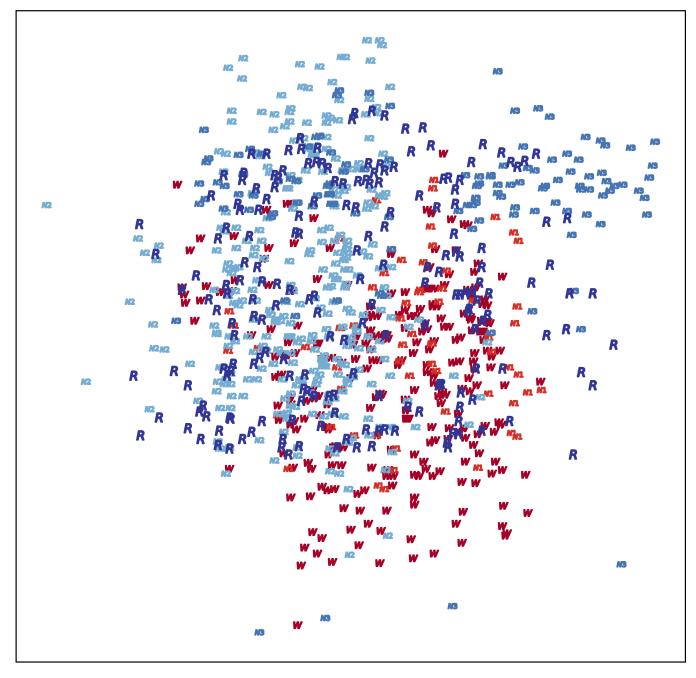}   %以pic.jpg的0.5倍大小输出
		\end{minipage}
	}
	\subfigure[After] %第二张子图
	{
		\begin{minipage}[t]{0.5\linewidth}
			\centering      %子图居中
			\includegraphics[scale=0.85]{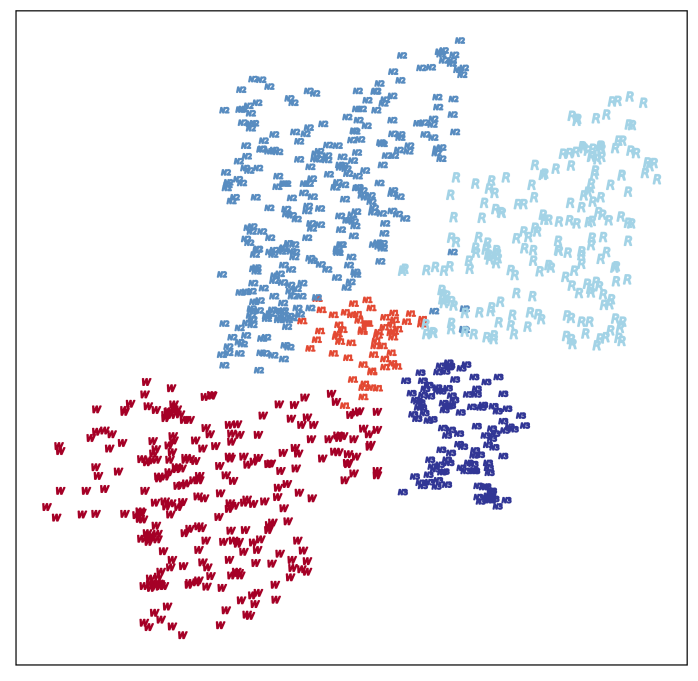}   %以pic.jpg的0.5倍大小输出
		\end{minipage}
	}
	
	\caption{The left panel shows the experimental results of the raw EEG and the right panel shows the results of the post-processing of the features extracted from the fully connected layer, where the different sleep stages are marked with different colours and labels, containing a total of 5 colours and 5 labels (W, N1, N2, N3, R).} %  %大图名称
\label{Figure11}
\end{figure}

To explore the effectiveness of the proposed model for sleep EEG feature extraction, we used the t-SNE (t-distributed stochastic neighbour embedding) method \cite{van2008visualizing} to visualize the distribution of EEG features before and after the model processing. We selected one subject containing 930 epochs and visualized the feature data obtained by processing the raw sleep EEG data and the fully connected layer of the model separately. %The results obtained are shown in Figure~\ref{Figure11}. 
According to the Fig.~\ref{Figure11}, without applying the new model, the sleep stages cannot be well distinguished with each other.% However, the same sleep stages were divided into the same regions and the boundaries appeared between different sleep stages, i.e., the differences between different sleep stages were significantly enhanced after our model extracting features. Moreover, the N1 stage and the R, N2 and W stages are all partially overlapped, which is consistent with the results of our previous experiments.

\section{Conclusion }
In this paper, the sleep staging method based on the dual attention network is proposed to address the problems of complex signal acquisition and poor staging in the current EEG-based automatic sleep staging method. The network can automatically extract valid information from the raw single-channel EEG and classify the sleep stages, realizing end-to-end learning. The network maintains good performance under different datasets and validation methods. The results show that the model satisfies the requirements of the manual sleep staging criteria and the overall performance of the method is better than current mainstream methods at the N1 stage.

Although the overall performance of the model in this paper has improved at stage N1 compared to other models, achieving an F1 score of 52.49, the accuracy is still low at 42.57. For this problem, a method that balances the proportion of sleep stages in the dataset and increases the weight of the loss function at stage N1 might improve the existing results to achieve higher accuracy. However, this deduction needs further validation to examine. In addition, without considering the arithmetic power, deepening the network constructed in this paper further can learn the features more completely, which can theoretically improve the network robustness as well as the overall performance.

\section{Acknowledgements}

This work was supported in part by the Beijing Municipal Education Commission Scientific Research Program under Grant KM202110009001 and the 2020 Hebei Provincial Science and Technology Plan Project under Grant 203777116D.


%% References


\bibliographystyle{model1-num-names}
\bibliography{sample.bib}

\begin{thebibliography}{37}
\expandafter\ifx\csname natexlab\endcsname\relax\def\natexlab#1{#1}\fi
\providecommand{\bibinfo}[2]{#2}
\ifx\xfnm\relax \def\xfnm[#1]{\unskip,\space#1}\fi
%Type = Article
\bibitem[{Laposky et~al.(2008)Laposky, Bass, Kohsaka, and
  Turek}]{laposky_bass_kohsaka_turek_2008}
\bibinfo{author}{A.~D. Laposky}, \bibinfo{author}{J.~Bass},
  \bibinfo{author}{A.~Kohsaka}, \bibinfo{author}{F.~W. Turek},
\newblock \bibinfo{title}{Sleep and circadian rhythms: Key components in the
  regulation of energy metabolism},
\newblock \bibinfo{journal}{FEBS Letters} \bibinfo{volume}{582}
  (\bibinfo{year}{2008}) \bibinfo{pages}{142–151}.
%Type = Article
\bibitem[{Stranges et~al.(2012)Stranges, Tigbe, G{\'o}mez-Oliv{\'e}, Thorogood,
  and Kandala}]{stranges2012sleep}
\bibinfo{author}{S.~Stranges}, \bibinfo{author}{W.~Tigbe},
  \bibinfo{author}{F.~X. G{\'o}mez-Oliv{\'e}}, \bibinfo{author}{M.~Thorogood},
  \bibinfo{author}{N.-B. Kandala},
\newblock \bibinfo{title}{Sleep problems: an emerging global epidemic? findings
  from the indepth who-sage study among more than 40,000 older adults from 8
  countries across africa and asia},
\newblock \bibinfo{journal}{Sleep} \bibinfo{volume}{35} (\bibinfo{year}{2012})
  \bibinfo{pages}{1173--1181}.
%Type = Article
\bibitem[{Imtiaz(2021)}]{imtiaz2021systematic}
\bibinfo{author}{S.~A. Imtiaz},
\newblock \bibinfo{title}{A systematic review of sensing technologies for
  wearable sleep staging},
\newblock \bibinfo{journal}{Sensors} \bibinfo{volume}{21}
  (\bibinfo{year}{2021}) \bibinfo{pages}{1562}.
%Type = Article
\bibitem[{Cho and Duffy(2018)}]{cho2018sleep}
\bibinfo{author}{J.~W. Cho}, \bibinfo{author}{J.~F. Duffy},
\newblock \bibinfo{title}{Sleep, sleep disorders, and sexual dysfunction.},
\newblock \bibinfo{journal}{The world journal of men's health}
  \bibinfo{volume}{37} (\bibinfo{year}{2018}) \bibinfo{pages}{261--275}.
%Type = Article
\bibitem[{Wolpert(1969)}]{wolpert1969manual}
\bibinfo{author}{E.~A. Wolpert},
\newblock \bibinfo{title}{A manual of standardized terminology, techniques and
  scoring system for sleep stages of human subjects.},
\newblock \bibinfo{journal}{Archives of General Psychiatry}
  \bibinfo{volume}{20} (\bibinfo{year}{1969}) \bibinfo{pages}{246--247}.
%Type = Article
\bibitem[{Berry et~al.(2012)Berry, Brooks, Gamaldo, Harding, Marcus, Vaughn
  et~al.}]{berry2012aasm}
\bibinfo{author}{R.~B. Berry}, \bibinfo{author}{R.~Brooks},
  \bibinfo{author}{C.~E. Gamaldo}, \bibinfo{author}{S.~M. Harding},
  \bibinfo{author}{C.~Marcus}, \bibinfo{author}{B.~V. Vaughn}, et~al.,
\newblock \bibinfo{title}{The aasm manual for the scoring of sleep and
  associated events},
\newblock \bibinfo{journal}{Rules, Terminology and Technical Specifications,
  Darien, Illinois, American Academy of Sleep Medicine} \bibinfo{volume}{176}
  (\bibinfo{year}{2012}) \bibinfo{pages}{2012}.
%Type = Article
\bibitem[{Chriskos et~al.(2020)Chriskos, Frantzidis, Nday, Gkivogkli, Bamidis,
  and Kourtidou-Papadeli}]{chriskos2020review}
\bibinfo{author}{P.~Chriskos}, \bibinfo{author}{C.~A. Frantzidis},
  \bibinfo{author}{C.~M. Nday}, \bibinfo{author}{P.~T. Gkivogkli},
  \bibinfo{author}{P.~D. Bamidis}, \bibinfo{author}{C.~Kourtidou-Papadeli},
\newblock \bibinfo{title}{A review on current trends in automatic sleep staging
  through bio-signal recordings and future challenges},
\newblock \bibinfo{journal}{Sleep Medicine Reviews}  (\bibinfo{year}{2020})
  \bibinfo{pages}{101377}.
%Type = Article
\bibitem[{Memar and Faradji(2017)}]{memar2017novel}
\bibinfo{author}{P.~Memar}, \bibinfo{author}{F.~Faradji},
\newblock \bibinfo{title}{A novel multi-class eeg-based sleep stage
  classification system},
\newblock \bibinfo{journal}{IEEE Transactions on Neural Systems and
  Rehabilitation Engineering} \bibinfo{volume}{26} (\bibinfo{year}{2017})
  \bibinfo{pages}{84--95}.
%Type = Article
\bibitem[{Tian et~al.(2017)Tian, Hu, Qi, Ye, Che, Ding, and
  Peng}]{tian2017hierarchical}
\bibinfo{author}{P.~Tian}, \bibinfo{author}{J.~Hu}, \bibinfo{author}{J.~Qi},
  \bibinfo{author}{X.~Ye}, \bibinfo{author}{D.~Che}, \bibinfo{author}{Y.~Ding},
  \bibinfo{author}{Y.~Peng},
\newblock \bibinfo{title}{A hierarchical classification method for automatic
  sleep scoring using multiscale entropy features and proportion information of
  sleep architecture},
\newblock \bibinfo{journal}{Biocybernetics and Biomedical Engineering}
  \bibinfo{volume}{37} (\bibinfo{year}{2017}) \bibinfo{pages}{263--271}.
%Type = Article
\bibitem[{Lajnef et~al.(2015)Lajnef, Chaibi, Ruby, Aguera, Eichenlaub, Samet,
  Kachouri, and Jerbi}]{lajnef2015learning}
\bibinfo{author}{T.~Lajnef}, \bibinfo{author}{S.~Chaibi},
  \bibinfo{author}{P.~Ruby}, \bibinfo{author}{P.-E. Aguera},
  \bibinfo{author}{J.-B. Eichenlaub}, \bibinfo{author}{M.~Samet},
  \bibinfo{author}{A.~Kachouri}, \bibinfo{author}{K.~Jerbi},
\newblock \bibinfo{title}{Learning machines and sleeping brains: automatic
  sleep stage classification using decision-tree multi-class support vector
  machines},
\newblock \bibinfo{journal}{Journal of neuroscience methods}
  \bibinfo{volume}{250} (\bibinfo{year}{2015}) \bibinfo{pages}{94--105}.
%Type = Article
\bibitem[{Hassan and Bhuiyan(2017)}]{hassan2017automated}
\bibinfo{author}{A.~R. Hassan}, \bibinfo{author}{M.~I.~H. Bhuiyan},
\newblock \bibinfo{title}{Automated identification of sleep states from eeg
  signals by means of ensemble empirical mode decomposition and random under
  sampling boosting},
\newblock \bibinfo{journal}{Computer methods and programs in biomedicine}
  \bibinfo{volume}{140} (\bibinfo{year}{2017}) \bibinfo{pages}{201--210}.
%Type = Inproceedings
\bibitem[{Samiee et~al.(2015)Samiee, Kov{\'a}cs, Kiranyaz, Gabbouj, and
  Saramaki}]{samiee2015sleep}
\bibinfo{author}{K.~Samiee}, \bibinfo{author}{P.~Kov{\'a}cs},
  \bibinfo{author}{S.~Kiranyaz}, \bibinfo{author}{M.~Gabbouj},
  \bibinfo{author}{T.~Saramaki},
\newblock \bibinfo{title}{Sleep stage classification using sparse rational
  decomposition of single channel eeg records},
\newblock in: \bibinfo{booktitle}{2015 23rd European Signal Processing
  Conference (EUSIPCO)}, \bibinfo{organization}{IEEE}, pp.
  \bibinfo{pages}{1860--1864}.
%Type = Article
\bibitem[{Sharma et~al.(2017)Sharma, Pachori, and
  Upadhyay}]{sharma2017automatic}
\bibinfo{author}{R.~Sharma}, \bibinfo{author}{R.~B. Pachori},
  \bibinfo{author}{A.~Upadhyay},
\newblock \bibinfo{title}{Automatic sleep stages classification based on
  iterative filtering of electroencephalogram signals},
\newblock \bibinfo{journal}{Neural Computing and Applications}
  \bibinfo{volume}{28} (\bibinfo{year}{2017}) \bibinfo{pages}{2959--2978}.
%Type = Article
\bibitem[{Jiang et~al.(2019)Jiang, Lu, Yu, and Yuanyuan}]{jiang2019robust}
\bibinfo{author}{D.~Jiang}, \bibinfo{author}{Y.-n. Lu},
  \bibinfo{author}{M.~Yu}, \bibinfo{author}{W.~Yuanyuan},
\newblock \bibinfo{title}{Robust sleep stage classification with single-channel
  eeg signals using multimodal decomposition and hmm-based refinement},
\newblock \bibinfo{journal}{Expert Systems with Applications}
  \bibinfo{volume}{121} (\bibinfo{year}{2019}) \bibinfo{pages}{188--203}.
%Type = Article
\bibitem[{Azadian et~al.(2019)Azadian, Yousefi~Rezaii, and
  Meshgini}]{azadian2019exploiting}
\bibinfo{author}{B.~Azadian}, \bibinfo{author}{T.~Yousefi~Rezaii},
  \bibinfo{author}{S.~Meshgini},
\newblock \bibinfo{title}{Exploiting sparse representation for sleep stage
  classification using electroencephalogram signal},
\newblock \bibinfo{journal}{Advanced Signal Processing} \bibinfo{volume}{3}
  (\bibinfo{year}{2019}) \bibinfo{pages}{1--11}.
%Type = Article
\bibitem[{Jadhav et~al.(2020)Jadhav, Rajguru, Datta, and
  Mukhopadhyay}]{jadhav2020automatic}
\bibinfo{author}{P.~Jadhav}, \bibinfo{author}{G.~Rajguru},
  \bibinfo{author}{D.~Datta}, \bibinfo{author}{S.~Mukhopadhyay},
\newblock \bibinfo{title}{Automatic sleep stage classification using
  time--frequency images of cwt and transfer learning using convolution neural
  network},
\newblock \bibinfo{journal}{Biocybernetics and Biomedical Engineering}
  \bibinfo{volume}{40} (\bibinfo{year}{2020}) \bibinfo{pages}{494--504}.
%Type = Article
\bibitem[{Tsinalis et~al.(2016)Tsinalis, Matthews, Guo, and
  Zafeiriou}]{tsinalis2016automatic}
\bibinfo{author}{O.~Tsinalis}, \bibinfo{author}{P.~M. Matthews},
  \bibinfo{author}{Y.~Guo}, \bibinfo{author}{S.~Zafeiriou},
\newblock \bibinfo{title}{Automatic sleep stage scoring with single-channel eeg
  using convolutional neural networks},
\newblock \bibinfo{journal}{arXiv preprint arXiv:1610.01683}
  (\bibinfo{year}{2016}).
%Type = Inproceedings
\bibitem[{Vilamala et~al.(2017)Vilamala, Madsen, and Hansen}]{vilamala2017deep}
\bibinfo{author}{A.~Vilamala}, \bibinfo{author}{K.~H. Madsen},
  \bibinfo{author}{L.~K. Hansen},
\newblock \bibinfo{title}{Deep convolutional neural networks for interpretable
  analysis of eeg sleep stage scoring},
\newblock in: \bibinfo{booktitle}{2017 IEEE 27th International Workshop on
  Machine Learning for Signal Processing (MLSP)}, \bibinfo{organization}{IEEE},
  pp. \bibinfo{pages}{1--6}.
%Type = Article
\bibitem[{Yang et~al.(2021)Yang, Zhu, Liu, and Liu}]{yang2021single}
\bibinfo{author}{B.~Yang}, \bibinfo{author}{X.~Zhu}, \bibinfo{author}{Y.~Liu},
  \bibinfo{author}{H.~Liu},
\newblock \bibinfo{title}{A single-channel eeg based automatic sleep stage
  classification method leveraging deep one-dimensional convolutional neural
  network and hidden markov model},
\newblock \bibinfo{journal}{Biomedical Signal Processing and Control}
  \bibinfo{volume}{68} (\bibinfo{year}{2021}) \bibinfo{pages}{102581}.
%Type = Article
\bibitem[{Seo et~al.(2020)Seo, Back, Lee, Park, Kim, and Lee}]{seo2020intra}
\bibinfo{author}{H.~Seo}, \bibinfo{author}{S.~Back}, \bibinfo{author}{S.~Lee},
  \bibinfo{author}{D.~Park}, \bibinfo{author}{T.~Kim},
  \bibinfo{author}{K.~Lee},
\newblock \bibinfo{title}{Intra-and inter-epoch temporal context network
  (iitnet) using sub-epoch features for automatic sleep scoring on raw
  single-channel eeg},
\newblock \bibinfo{journal}{Biomedical Signal Processing and Control}
  \bibinfo{volume}{61} (\bibinfo{year}{2020}) \bibinfo{pages}{102037}.
%Type = Article
\bibitem[{Eldele et~al.(2021)Eldele, Chen, Liu, Wu, Kwoh, Li, and
  Guan}]{eldele2021attention}
\bibinfo{author}{E.~Eldele}, \bibinfo{author}{Z.~Chen},
  \bibinfo{author}{C.~Liu}, \bibinfo{author}{M.~Wu}, \bibinfo{author}{C.-K.
  Kwoh}, \bibinfo{author}{X.~Li}, \bibinfo{author}{C.~Guan},
\newblock \bibinfo{title}{An attention-based deep learning approach for sleep
  stage classification with single-channel eeg},
\newblock \bibinfo{journal}{IEEE Transactions on Neural Systems and
  Rehabilitation Engineering} \bibinfo{volume}{29} (\bibinfo{year}{2021})
  \bibinfo{pages}{809--818}.
%Type = Article
\bibitem[{Yildirim et~al.(2019)Yildirim, Baloglu, and
  Acharya}]{yildirim2019deep}
\bibinfo{author}{O.~Yildirim}, \bibinfo{author}{U.~B. Baloglu},
  \bibinfo{author}{U.~R. Acharya},
\newblock \bibinfo{title}{A deep learning model for automated sleep stages
  classification using psg signals},
\newblock \bibinfo{journal}{International journal of environmental research and
  public health} \bibinfo{volume}{16} (\bibinfo{year}{2019})
  \bibinfo{pages}{599}.
%Type = Inproceedings
\bibitem[{Humayun et~al.(2019)Humayun, Sushmit, Hasan, and
  Bhuiyan}]{humayun2019end}
\bibinfo{author}{A.~I. Humayun}, \bibinfo{author}{A.~S. Sushmit},
  \bibinfo{author}{T.~Hasan}, \bibinfo{author}{M.~I.~H. Bhuiyan},
\newblock \bibinfo{title}{End-to-end sleep staging with raw single channel eeg
  using deep residual convnets},
\newblock in: \bibinfo{booktitle}{2019 IEEE EMBS International Conference on
  Biomedical \& Health Informatics (BHI)}, \bibinfo{organization}{IEEE}, pp.
  \bibinfo{pages}{1--5}.
%Type = Inproceedings
\bibitem[{Olesen et~al.(2018)Olesen, Jennum, Peppard, Mignot, and
  Sorensen}]{olesen2018deep}
\bibinfo{author}{A.~N. Olesen}, \bibinfo{author}{P.~Jennum},
  \bibinfo{author}{P.~Peppard}, \bibinfo{author}{E.~Mignot},
  \bibinfo{author}{H.~B. Sorensen},
\newblock \bibinfo{title}{Deep residual networks for automatic sleep stage
  classification of raw polysomnographic waveforms},
\newblock in: \bibinfo{booktitle}{2018 40th Annual International Conference of
  the IEEE Engineering in Medicine and Biology Society (EMBC)},
  \bibinfo{organization}{IEEE}, pp. \bibinfo{pages}{1--4}.
%Type = Inproceedings
\bibitem[{Phan et~al.(2018)Phan, Andreotti, Cooray, Ch{\'e}n, and
  De~Vos}]{phan2018automatic}
\bibinfo{author}{H.~Phan}, \bibinfo{author}{F.~Andreotti},
  \bibinfo{author}{N.~Cooray}, \bibinfo{author}{O.~Y. Ch{\'e}n},
  \bibinfo{author}{M.~De~Vos},
\newblock \bibinfo{title}{Automatic sleep stage classification using
  single-channel eeg: Learning sequential features with attention-based
  recurrent neural networks},
\newblock in: \bibinfo{booktitle}{2018 40th Annual International Conference of
  the IEEE Engineering in Medicine and Biology Society (EMBC)},
  \bibinfo{organization}{IEEE}, pp. \bibinfo{pages}{1452--1455}.
%Type = Inproceedings
\bibitem[{He et~al.(2016)He, Zhang, Ren, and Sun}]{he2016deep}
\bibinfo{author}{K.~He}, \bibinfo{author}{X.~Zhang}, \bibinfo{author}{S.~Ren},
  \bibinfo{author}{J.~Sun},
\newblock \bibinfo{title}{Deep residual learning for image recognition},
\newblock in: \bibinfo{booktitle}{Proceedings of the IEEE conference on
  computer vision and pattern recognition}, pp. \bibinfo{pages}{770--778}.
%Type = Article
\bibitem[{Donoho(1995)}]{donoho1995noising}
\bibinfo{author}{D.~L. Donoho},
\newblock \bibinfo{title}{De-noising by soft-thresholding},
\newblock \bibinfo{journal}{IEEE transactions on information theory}
  \bibinfo{volume}{41} (\bibinfo{year}{1995}) \bibinfo{pages}{613--627}.
%Type = Article
\bibitem[{Zhao et~al.(2019)Zhao, Zhong, Fu, Tang, and Pecht}]{zhao2019deep}
\bibinfo{author}{M.~Zhao}, \bibinfo{author}{S.~Zhong}, \bibinfo{author}{X.~Fu},
  \bibinfo{author}{B.~Tang}, \bibinfo{author}{M.~Pecht},
\newblock \bibinfo{title}{Deep residual shrinkage networks for fault
  diagnosis},
\newblock \bibinfo{journal}{IEEE Transactions on Industrial Informatics}
  \bibinfo{volume}{16} (\bibinfo{year}{2019}) \bibinfo{pages}{4681--4690}.
%Type = Article
\bibitem[{Goldberger et~al.(2000)Goldberger, Amaral, Glass, Hausdorff, Ivanov,
  Mark, Mietus, Moody, Peng, and Stanley}]{goldberger2000physiobank}
\bibinfo{author}{A.~L. Goldberger}, \bibinfo{author}{L.~A. Amaral},
  \bibinfo{author}{L.~Glass}, \bibinfo{author}{J.~M. Hausdorff},
  \bibinfo{author}{P.~C. Ivanov}, \bibinfo{author}{R.~G. Mark},
  \bibinfo{author}{J.~E. Mietus}, \bibinfo{author}{G.~B. Moody},
  \bibinfo{author}{C.-K. Peng}, \bibinfo{author}{H.~E. Stanley},
\newblock \bibinfo{title}{Physiobank, physiotoolkit, and physionet: components
  of a new research resource for complex physiologic signals},
\newblock \bibinfo{journal}{circulation} \bibinfo{volume}{101}
  (\bibinfo{year}{2000}) \bibinfo{pages}{e215--e220}.
%Type = Article
\bibitem[{Kemp et~al.(2000)Kemp, Zwinderman, Tuk, Kamphuisen, and
  Oberye}]{kemp2000analysis}
\bibinfo{author}{B.~Kemp}, \bibinfo{author}{A.~H. Zwinderman},
  \bibinfo{author}{B.~Tuk}, \bibinfo{author}{H.~A. Kamphuisen},
  \bibinfo{author}{J.~J. Oberye},
\newblock \bibinfo{title}{Analysis of a sleep-dependent neuronal feedback loop:
  the slow-wave microcontinuity of the eeg},
\newblock \bibinfo{journal}{IEEE Transactions on Biomedical Engineering}
  \bibinfo{volume}{47} (\bibinfo{year}{2000}) \bibinfo{pages}{1185--1194}.
%Type = Article
\bibitem[{Perslev et~al.(2019)Perslev, Jensen, Darkner, Jennum, and
  Igel}]{perslev2019u}
\bibinfo{author}{M.~Perslev}, \bibinfo{author}{M.~H. Jensen},
  \bibinfo{author}{S.~Darkner}, \bibinfo{author}{P.~J. Jennum},
  \bibinfo{author}{C.~Igel},
\newblock \bibinfo{title}{U-time: A fully convolutional network for time series
  segmentation applied to sleep staging},
\newblock \bibinfo{journal}{arXiv preprint arXiv:1910.11162}
  (\bibinfo{year}{2019}).
%Type = Article
\bibitem[{Supratak et~al.(2017)Supratak, Dong, Wu, and
  Guo}]{supratak2017deepsleepnet}
\bibinfo{author}{A.~Supratak}, \bibinfo{author}{H.~Dong},
  \bibinfo{author}{C.~Wu}, \bibinfo{author}{Y.~Guo},
\newblock \bibinfo{title}{Deepsleepnet: A model for automatic sleep stage
  scoring based on raw single-channel eeg},
\newblock \bibinfo{journal}{IEEE Transactions on Neural Systems and
  Rehabilitation Engineering} \bibinfo{volume}{25} (\bibinfo{year}{2017})
  \bibinfo{pages}{1998--2008}.
%Type = Article
\bibitem[{Zhang et~al.(2016)Zhang, Wu, Bai, and Chen}]{zhang2016automatic}
\bibinfo{author}{J.~Zhang}, \bibinfo{author}{Y.~Wu}, \bibinfo{author}{J.~Bai},
  \bibinfo{author}{F.~Chen},
\newblock \bibinfo{title}{Automatic sleep stage classification based on sparse
  deep belief net and combination of multiple classifiers},
\newblock \bibinfo{journal}{Transactions of the Institute of Measurement and
  Control} \bibinfo{volume}{38} (\bibinfo{year}{2016})
  \bibinfo{pages}{435--451}.
%Type = Article
\bibitem[{Phan et~al.(2020)Phan, Ch{\'e}n, Koch, Lu, McLoughlin, Mertins, and
  De~Vos}]{phan2020towards}
\bibinfo{author}{H.~Phan}, \bibinfo{author}{O.~Y. Ch{\'e}n},
  \bibinfo{author}{P.~Koch}, \bibinfo{author}{Z.~Lu},
  \bibinfo{author}{I.~McLoughlin}, \bibinfo{author}{A.~Mertins},
  \bibinfo{author}{M.~De~Vos},
\newblock \bibinfo{title}{Towards more accurate automatic sleep staging via
  deep transfer learning},
\newblock \bibinfo{journal}{IEEE Transactions on Biomedical Engineering}
  (\bibinfo{year}{2020}).
%Type = Article
\bibitem[{Phan et~al.(2018)Phan, Andreotti, Cooray, Ch{\'e}n, and
  De~Vos}]{phan2018joint}
\bibinfo{author}{H.~Phan}, \bibinfo{author}{F.~Andreotti},
  \bibinfo{author}{N.~Cooray}, \bibinfo{author}{O.~Y. Ch{\'e}n},
  \bibinfo{author}{M.~De~Vos},
\newblock \bibinfo{title}{Joint classification and prediction cnn framework for
  automatic sleep stage classification},
\newblock \bibinfo{journal}{IEEE Transactions on Biomedical Engineering}
  \bibinfo{volume}{66} (\bibinfo{year}{2018}) \bibinfo{pages}{1285--1296}.
%Type = Article
\bibitem[{Silber et~al.(2007)Silber, Ancoli-Israel, Bonnet, Chokroverty,
  Grigg-Damberger, Hirshkowitz, Kapen, Keenan, Kryger, Penzel
  et~al.}]{silber2007visual}
\bibinfo{author}{M.~H. Silber}, \bibinfo{author}{S.~Ancoli-Israel},
  \bibinfo{author}{M.~H. Bonnet}, \bibinfo{author}{S.~Chokroverty},
  \bibinfo{author}{M.~M. Grigg-Damberger}, \bibinfo{author}{M.~Hirshkowitz},
  \bibinfo{author}{S.~Kapen}, \bibinfo{author}{S.~A. Keenan},
  \bibinfo{author}{M.~H. Kryger}, \bibinfo{author}{T.~Penzel}, et~al.,
\newblock \bibinfo{title}{The visual scoring of sleep in adults},
\newblock \bibinfo{journal}{Journal of clinical sleep medicine}
  \bibinfo{volume}{3} (\bibinfo{year}{2007}) \bibinfo{pages}{121--131}.
%Type = Article
\bibitem[{Van~der Maaten and Hinton(2008)}]{van2008visualizing}
\bibinfo{author}{L.~Van~der Maaten}, \bibinfo{author}{G.~Hinton},
\newblock \bibinfo{title}{Visualizing data using t-sne.},
\newblock \bibinfo{journal}{Journal of machine learning research}
  \bibinfo{volume}{9} (\bibinfo{year}{2008}).

\end{thebibliography}






% \end{thebibliography}

\end{document}